\documentclass[10pt,journal,compsoc]{IEEEtran}

\ifCLASSOPTIONcompsoc
  \usepackage[nocompress]{cite}
\else
  \usepackage{cite}
\fi

\usepackage{graphicx}
\usepackage{booktabs} 
\usepackage{amsmath}
\usepackage{hyperref}
\usepackage{xcolor}
\usepackage{multirow}
\usepackage{subcaption}
\usepackage{changes}
\usepackage{amsmath}
\usepackage{diagbox}
\definechangesauthor[name={Iztok Fister}, color=red]{iztok}

\usepackage{algorithm}
\usepackage{algpseudocode}

\algnewcommand\algorithmicforeach{\textbf{for each}}
\algdef{S}[FOR]{ForEach}[1]{\algorithmicforeach\ #1\ \algorithmicdo}

\usepackage{enumitem}

\newlist{RQ}{enumerate}{1}
\setlist[RQ]{label=RQ-\arabic*:}

\usepackage{array,booktabs,ragged2e}
\newcolumntype{R}[1]{>{\RaggedRight\arraybackslash}p{#1}}
\newcolumntype{C}[1]{>{\centering\arraybackslash}p{#1}}
\newcolumntype{L}[1]{>{\RaggedLeft\arraybackslash}p{#1}}

\ifCLASSINFOpdf
\else
\fi

\begin{document}

\title{NiaAutoARM: Automated generation and evaluation of Association Rule Mining pipelines}

\author{Uro\v{s} Mlakar,~\IEEEmembership{Member,~IEEE,}
        Iztok Fister Jr.,~\IEEEmembership{Member,~IEEE,}
        Iztok Fister,~\IEEEmembership{Member,~IEEE}

\IEEEcompsocitemizethanks{\IEEEcompsocthanksitem Uro\v{s} Mlakar, Iztok Fister Jr. and Iztok Fister are with the Faculty of Electrical Engineering and Computer Science, University of Maribor, Koro\v{s}ka cesta 46, 2000 Maribor, Slovenia. Email: iztok.fister1@um.si\protect\\}}

\IEEEtitleabstractindextext{
\begin{abstract}
The Numerical Association Rule Mining paradigm that includes concurrent dealing with numerical and categorical attributes is beneficial for discovering associations from datasets consisting of both features. The process is not considered as easy since it incorporates several processing steps running sequentially that form an entire pipeline, e.g., preprocessing, algorithm selection, hyper-parameter optimization, and the definition of metrics evaluating the quality of the association rule. In this paper, we proposed a novel Automated Machine Learning method, NiaAutoARM, for constructing the full association rule mining pipelines based on stochastic population-based meta-heuristics automatically. Along with the theoretical representation of the proposed method, we also present a comprehensive experimental evaluation of the proposed method.
\end{abstract}

}

\maketitle

\IEEEdisplaynontitleabstractindextext

\IEEEpeerreviewmaketitle

\IEEEraisesectionheading{\section{Introduction}\label{sec:introduction}}
\IEEEPARstart{T}{he} design of Machine Learning (ML) pipelines usually demands user interaction to select appropriate preprocessing methods, perform feature engineering, select the most appropriate ML method, and set a combination of hyper-parameters~\cite{yao2018taking}. Therefore, preparing an ML pipeline is complex, and, primarily, inappropriate for non-specialists in the Data Science or Artificial Intelligence domains~\cite{hutterautomated}. On the other hand, tuning the entire pipeline to produce the best results also may involve a lot of time for the users, mainly if we deal with very complex datasets. 

Automated Machine Learning (AutoML) methods have been appeared, to draw the application of ML methods nearer to the users in the sense of ML democratization~\cite{hutterautomated,he2021automl}. The main benefit of these methods is searching for the best pipeline in different ML tasks automatically. Until recently, AutoML forms can be found for solving classification problems, neural architecture search, regression problems~\cite{conrad2022benchmarking}, and reinforcement learning.  

Association Rule Mining (ARM) is a ML method for discovering the relationships between items in transaction databases. Bare ARM is limited, since it operates initially with categorical type of attributes only. Recently, Numerical Association Rule Mining (NARM) has been proposed, that is a variant of a bare ARM, which allows dealing with numerical and categorical attributes concurrently, and thus removes the bottleneck of the bare ARM. The NARM also delivers several benefits, since the results can be more reliable and accurate, and contain less noise than bare ARM, where the numerical attributes need to be discretized before use. Nowadays, the problem of NARM is tackled mainly by using population-based meta-heuristics, which can cope large search spaces effectively. Let us mention that the acronym ARM is used as synonym for the acronym NARM in the paper.

\begin{figure*}[htb]
\centering
\includegraphics[width=16cm]{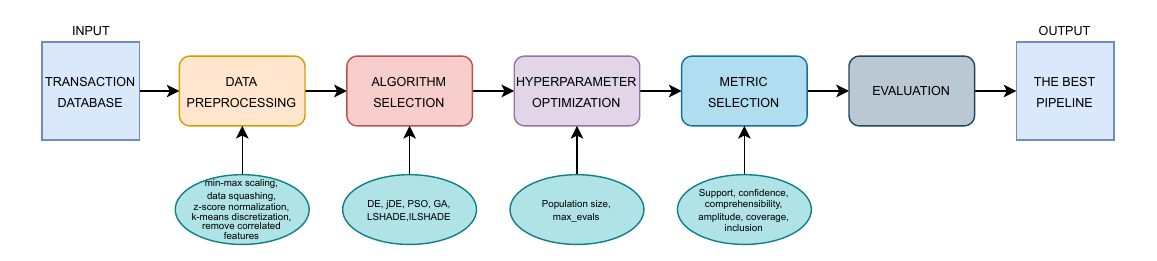}
\caption{The structure of the basic ARM pipeline.}
\label{basic-pipeline}
\end{figure*}

The ARM pipeline (see Fig.~\ref{basic-pipeline}) is far from being uncomplicated, since it consists of several components, as follows: (1) data preprocessing, (2) mining algorithm selection, (3) hyper-parameter optimization, (4) evaluation metric selection, and (5) evaluation. Each of these components can be implemented using several ML methods.
Consequently, composing the ARM pipeline manually requires a lot of human intervention, potentially a time-consuming task. Therefore, automation of this composing led us to the new domain of AutoML, i.e., AutoARM.

The data entering the ARM  pipeline are in the form of a transaction database; the optional first component of the ARM pipeline is preprocessing, where the data can be preprocessed further using various ML methods. The selection of the proper processing component presents a crucial step, where the most appropriate population-based meta-heuristic Nature-Inspired Algorithm (Nia) needs to be determined for ARM. Mainly, the NI algorithms encompasses two classes of population-based algorithms: Evolutionary Algorithms (EA)~\cite{eiben2015introduction} and Swarm Intelligence (SI) based~\cite{blum2008swarm}.

According to previous studies, no universal population-based meta-heuristic exists for ARM achieving the best results by mining all datasets. This phenomenon is also justified by the No Free Lunch (NFL) theorem of Wolpert and Macready~\cite{wolpert1997no}. The next component in the pipeline is the hyper-parameter optimization for the selected population-based meta-heuristic, where the best combination of hyper-parameters is searched for. Finally, the selection of the favorable association rules depends on the composition of the more suitable metrics captured in the fitness function. In our case, the fitness function is represented as a linear combination of several ARM metrics (e.g., support, confidence, amplitude, etc.) weighted with particular weights. 

To the best knowledge of the authors, no specific AutoML methods exist for constructing the ARM pipelines automatically. Therefore, the contributions of this study are:
\begin{itemize}
\item To propose the first AutoML solution for searching for the best ARM pipeline, where this automatic searching is represented as an optimization problem.
\item To dedicate special attention to the preprocessing steps of ARM, which have been neglected slightly in the recent research works,
\item To implement a new method called NiaAutoARM as a Python package,
\item To evaluate the proposed method rigorously on several datasets.
\end{itemize}

The structure of the remainder of the paper is as follows: The materials and methods, needed for understanding the observed subjects that follow, are discussed in Section~\ref{sec:related}. The proposed method for automated ARM is described in Section~\ref{sec:niaautoarm} in detail. The experiments and the obtained results are the subjects of Section~\ref{sec:experiments}. A short discussion of the results is presented in Section~\ref{sec:discussion}. The paper is concluded in Section~\ref{sec:conclusion}, with a summarization of the performed work, and outlining the potential directions for the future work.

\section{Related work}\label{sec:related}
The section highlights topics necessary for understanding the subjects of the paper. In line with this, the following topics are dealt with:
\begin{itemize}
    \item AutoML,
    \item NiaAML,
    \item NiaAutoARM.
 \end{itemize}
The mentioned topics are discussed in detail in the remainder of the paper.

\subsection{AutoML}
Using ML methods in practice demands experienced human ML experts, who are typically expensive and hard to find on the market. On the other hand, computing has become cheaper day by day. This fact has led to the advent of AutoML, which is capable of constructing the ML pipelines of a similar, or even better quality, than by the human experts~\cite{hutterautomated}. Consequently, the AutoML enables the so-called democratization of ML. This means that the usage of the ML methods is drawn closer to the user by AutoML, and, thus, this technology tries to avoid the principle of user-in-the-loop~\cite{holzinger2024human}.

Automation of ML methods is allowed by AutoML using the ML pipelines. Indeed, these pipelines are the control points of the AutoML system. Typically, the ML pipeline consists of the following processing steps:
\begin{itemize}
\item preprocessing,
\item processing with definite ML methods,
\item hyper-parameter optimization,
\item evaluation.
\end{itemize}

AutoML is, nowadays, a very studied research area. The recent advances in the field have been summarized in several review papers~\cite{he2021automl,zoller2021benchmark,yao2018taking,escalante2021automated}. There also exist a dozen applications of AutoML~\cite{musigmann2022testing,barreiro2018net}, where the special position is devoted to NiaAML, which is discussed in more detail in the remainder of this section.

\subsection{NiaAML}
The NiaAML is an AutoML method based on stochastic Nia-s for optimization, where the AutoML is modeled as an optimization problem. The first version of NiaAML~\cite{niaaml1} covers composing classification pipelines, where a stochastic Nia searches for the best classification pipeline. The following steps are included in the AutoML pipeline, i.e., automatic feature selection, feature scaling, classifier selection, and hyper-parameter optimization. Each classifier configuration found by the optimizers is tested using cross-validation. 

Following the NiaAML, the NiaAML2~\cite{niaaml2,niaaml1} was proposed, which eliminated the main weakness of the original NiaAML method, where the hyper-parameters' optimization is performed simultaneously with the construction of the classification pipelines in a single phase. In the NiaAML, only one instance of the stochastic algorithm was needed. However, in the NiaAML2, the construction of the pipeline and hyper-parameter optimization was divided into two separate phases, where two instances of nature-inspired algorithms were deployed, one after the other, to cover both steps.  The first step covers the composition of the classification of the pipeline, while the second is devoted to hyper-parameter optimization.

\subsection{NiaARM}
The NiaARM is a Python framework~\cite{stupan2022niaarm} that implements the ARM-DE algorithm comprehensively~\cite{fister2018differential}, where the ARM is modeled as a
single objective, continuous optimization problem. The fitness function in NiaARM is defined as a weighted sum of arbitrary evaluation metrics. One of the most vital points of NiaARM is that it is based on the NiaPy framework~\cite{NiaPyJOSS2018}, and, thus, different Nia-s can be used in the optimizer role. According to the knowledge of the authors, NiaARM is the only comprehensive framework for NARM where all NARM steps are implemented, i.e., preprocessing, optimization, and visualization. Other benefits of NiaARM are good documentation and many examples provided by the maintainers, Command Line Interface (CLI), are easy to use.

\section{Proposed method: NiaAutoARM}\label{sec:niaautoarm}
The proposed method NiaAutoARM is inspired mainly by the NiaAML method. Thus, we define the problem of ARM pipeline construction as a continuous optimization problem. This means that an arbitrary population-based meta-heuristic Nia, which works in a continuous search space, can be applied for solving this problem. Indeed, the NiaAutoARM works as an outer layered meta-heuristic, that controls the behavior of the inner layered NI heuristic for ARM by searching for the optimal inner algorithm, the corresponding hyper-parameters, the employed preprocessing methods, and the outline of the proper evaluation function.  

In the NiaAutoARM, each individual in the population of solutions represents one feasible ARM pipeline as:
\begin{equation} 
\label{eq:1}
\centering
\small
\begin{split}
\mathbf{x}_{i}^{(t)}=&\left\langle\underbrace{x_{i, 1}^{(t)}}_{\text{ALGORITHM}}, \underbrace{y_{i, 1}^{(t)},y_{i, 2}^{(t)}}_{\text{HYPER-PARAMETERS}}, \underbrace{p_{i, 1}^{(t)},\ldots, p_{i, P}^{(t)}}_{\text{PREPROCESSING}}, \right .\\
&\left . \underbrace{z_{i, 1}^{(t)},\ldots,z_{i, M}^{(t)}}_{\text{METRICS}}, \underbrace{w_{i, 1}^{(t)}, \ldots, w_{i, M}^{(t)}}_{\text{METRIC WEIGHTS}}\right\rangle,
\end{split}
\end{equation}
where parameter $P$ denotes the number of potential preprocessing methods, and parameter $M$ is the number of potential ARM metrics to be applied. As is evident from Eq.~(\ref{eq:1}), each real-valued element of solution in a genotype search space within the interval $[0,1]$ encodes the particular NiaAutoARM component of the pipeline in a phenotype solution space as follows: The $\text{ALGORITHM}$ component denotes the stochastic population-based Nia, which is chosen from the pool of available algorithms, typically selected by the user from a NiaPy library relatively to the value of $x^{(t)}_{i,1}$~\cite{stupan2022niaarm}. 

The $\text{HYPER-PARAMETERS}$ component indicates a magnitude of two parameters: the maximum number of individuals $\mathit{NP}$, and the maximum number of fitness function evaluations $\mathit{MAXFES}$ as a termination condition for the selected algorithm. Both values, $y_{i, 1}^{(t)}$ and $y_{i, 2}^{(t)}$, are mapped in genotype-phenotype mapping to the specific domain of the mentioned parameters as proposed by Mlakar et al. in~\cite{mlakar2023variable}. The $\text{PREPROCESSING}$ component determines the pool of available preprocessing algorithms which can be applied to the dataset. On the one hand, if $P=0$, no preprocessing algorithm is applied, while, on the other hand, if $P > 0$ and $p_{i, j}^{(t)}>.5$ for $j=1,\ldots,P$, the $j$-th preprocessing algorithms from the pool will be observed for applying to the dataset. The $\text{METRICS}$ component is reserved for the pool of $M$ rule evaluation metrics devoted for estimating the quality of the mined association rules. Additionally, the weights of the metrics are included by the $METRIC\_WEIGHTS$ component, which weighs the influence of the particular evaluation metric on the appropriate association rule. Typically, the evaluation metrics as illustrated in Table~\ref{tab:evaluation_metrics} are employed in NiaAutoARM.
\begin{table}[!h]
    \centering
    \caption{ARM metrics used for evaluating the mined rules.}
    \label{tab:evaluation_metrics}
    \begin{tabular}{l|l}
        Metric &Evaluation functions\\
        \hline
        Support &  $Supp(X \implies Y) = \frac{|{t_i|t_i \in X \wedge t_i \in Y}|}{N}$\\
        Confidence &  $Conf(X \implies Y) = \frac{Supp(X \cup Y)}{Supp(X)}$\\
        Coverage & $Cover(X \implies Y) = \frac{|{t_i|t_i \in Y}|}{M}$\\
        Amplitude & $Amp(X \implies Y) = \frac{Supp(X \cap Y)}{Supp(X)} - \frac{Supp(Y)}{N}$\\
        Inclusion & $Incl(X \implies Y) = \frac{Supp(X \cap Y)}{Supp(X)}$\\
        Comprehensibility & $Comp(X \implies Y) = \frac{Supp(X \cap Y)}{Supp(Y)}$\\
        \hline
    \end{tabular}
\end{table}

\begin{figure*}[!ht]
    \centering
    \includegraphics{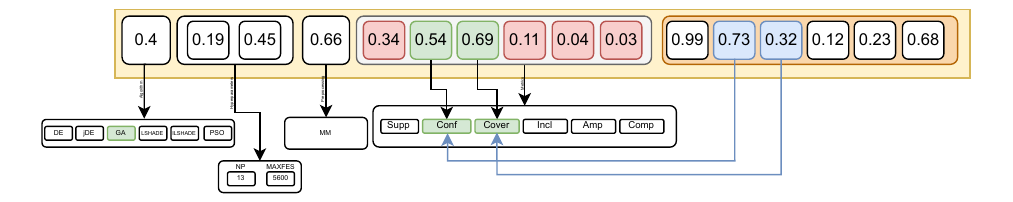}
    \caption{An example of genotype-phenotype mapping within the ARM pipeline construction.} 
    \label{fig:decode_pipeline}
\end{figure*}

An example of decoding an ARM pipeline to the solution space is illustrated in Figure~\ref{fig:decode_pipeline}, where the parameters are set as: $P=1$ and $M=6$. Let us suppose that the pool of inner algorithms is given as $\mathbf{Alg}=(\text{PSO,DE,GA,iLShade, LShade,jDE})$ denoting Differential Evolution (DE)~\cite{storn1997differential}, Particle Swarm Optimization (PSO)~\cite{kennedy1995particle}, Genetic Algorithm (GA)~\cite{Goldberg1989genetic}, Success-History based Adaptive DE with Linear size reduction (LSHADE)~\cite{tanabe2014improving}, Improved LSHADE (ILSHADE)~\cite{Brest2016il-shade}, and self-adaptive DE (jDE)~\cite{brest2006self}. Furthermore, the domains of the hyper-parameters are set to $\mathit{NP}\in[10,30]$ and $\mathit{MAXFES}\in[2000,10000]$. Moreover, the pool of preprocessing methods is defined as $\mathbf{Prep}=(\text{MM,ZS,DS,RHC,DK})$ designating "Min\_Max normalization" (MM), "Z-Score normalization" (ZS), "Data Squashing" (DS), "Remove Highly Correlated features" (RHC) and "Discretization K-means" (DK), while the pool of ARM metrics and weights as $\mathbf{Metr}=(\text{Supp,Conf,Cover,Amp,Incl,Comp})$, where elements are referred to the evaluation functions in Table~\ref{tab:evaluation_metrics}.

As a result, the outer meta-heuristic algorithm calls the inner heuristic algorithm as follows:
\begin{scriptsize}
\begin{equation}
    \underbrace{*Alg[\Gamma(x_{i,1})]}_{\text{Algorithm call}}(\underbrace{P,M}_{\text{Param}},\underbrace{\Gamma(y_{i,1})}_{\text{NP}},\underbrace{\Gamma(y_{i,2})}_{\text{MAXFES}},\underbrace{\Gamma(\textbf{Prep},\textbf{p})}_{\text{Preprocess}},\underbrace{\Gamma(\textbf{Metr},\textbf{z})}_{\text{Metrics}}),\underbrace{\Gamma(\textbf{Metr},\textbf{w})}_{\text{Weights}})),
\end{equation}
\end{scriptsize}
where the function $\Gamma$ denotes the mapping of genotype values to the phenotype values. Let us mention that the scalar values of 'Algorithm call', NP and MAXFES are decoded by mapping their values from the interval [0,1] to the domain values in the solution space. On the other hand, the preprocessing methods and ARM metrics represent sets, where each member is taken from the sets $\mathbf{Prep}$ and $\mathbf{Matr}$ according to the probability 0.5, based on the values of the vectors $\mathbf{p}$ and $\mathbf{z}$. Interestingly, the weight vector can be treated either statically or adaptively w.r.t., setting a parameter $\mathit{weight\_adaptation}$.  When the parameter is set as \textbf{true},  the adapted values from vector $\mathbf{w}$ indicate an impact of a definite ARM metric in the linear combination of ARM metrics within the fitness function. If this parameter is set to \textbf{false}, the values are fixed to the value 1.0.

Although the quality of the mined association rules is calculated in the inner algorithm using the weighted linear combination of the ARM metrics, the NiaAutoARM estimates the quality of the pipeline due to the fairness using the fitness function as:
\begin{equation}
    f(\textbf{x}_i^{(t)}) = \frac{\alpha\cdot \mathit{supp}(X\implies Y)+\beta\cdot \mathit{conf}(X\implies Y)}{\alpha+\beta},
    \label{eq:surrogate_fitness}
\end{equation}
where $\alpha$ and $\beta$ designate the impact of the definite ARM metric on the quality of the solution. It is discarded, if no rules are produced or the pipeline fails to decode to the solution space.

The pseudo-code of the proposed NiaAutoARM for constructing the classification pipelines is presented in Algorithm~\ref{alg:niaaml},
\begin{algorithm}[htb]
\caption{A pseudo-code of the NiaAutoARM method.}
\label{alg:niaaml}
\small
\begin{algorithmic}[1]
\State $\mathbf{P}\leftarrow$ \Call{Initialize\_real-valued\_vectors\_randomly}{$\mathbf{x}_i$}
\State $best\_pipeline \leftarrow$ \Call{Eval\_and\_select\_the\_best} {$\mathbf{P}$}
\While {\textsc{Termination\_condition\_not\_met}}
    \ForEach {$\mathbf{x}_i \in \mathbf{P}$}
        \State $\mathbf{x}_{trial} \leftarrow$ \Call{Modify\_using\_NI\_algorithm}{$\mathbf{x}_i$}
        \State $pipeline \leftarrow$ \Call{Construct\_pipeline}{$\mathbf{x}_{trial}$}
        \State $cur\_pipeline \leftarrow$ \Call{Construct\_pipeline}{$\mathbf{x}_{i}$}
        \If {\Call{Eval}{$pipeline$} $\geq$ \Call{Eval}{$cur\_pipeline$}}
            \State $\mathbf{x}_i \leftarrow \mathbf{x}_{trial}$\quad\Comment{Replace the worse individual}
        \EndIf
        \If {\Call{Eval}{$pipeline$} $\geq$ \Call{Eval}{$best\_pipeline$}}
            \State $best\_pipeline \leftarrow pipeline$
        \EndIf
    \EndFor
\EndWhile
\State {\Return {$best\_pipeline$}}
\end{algorithmic}
\normalsize
\end{algorithm}
from which it can be observed that the outer meta-heuristic starts with a random initialization of the population (function \textsc{Initialize\_real-valued\_vectors\_randomly} in line~1). After evaluation regarding Eq.~(\ref{eq:surrogate_fitness}) and determining the best solution (function \textsc{Eval\_and\_select\_the\_best in line~2}), the evolution cycle is started (lines 3-15) that is terminated using function \textsc{Termination\_condition\_not\_met}. Within the evolution cycle, each individual $\mathbf{x}_i$ in the population $\mathbf{P}$ (lines 4-14) is at first modified (function \textsc{Modify\_using\_NI\_algorithms} in line 5). This modification results in the production of a trial solution $\mathbf{x}_{trial}$. Next, both the trial and target solutions are mapped to the phenotype solution space producing the trial $\mathit{pipeline}$ and target $\mathit{cur\_pipeline}$ (also the current best) solutions (lines~6 and~7). If the fitness function value of the trial pipeline is better that of the current best evaluated using $\textsc{Eval}$ function (line 8), the target solution becomes trial (line~9). Finally, if the trial pipeline is even better than the global best pipeline, $\textit{best\_pipeline}$ (line~11), the global best pipeline becomes the trial pipeline (line~12). 

\section{Experimental evaluation}\label{sec:experiments}

The primary goal of the experiments was to evaluate whether NiaAutoARM can find an optimal pipeline for solving various ARM problems automatically. A series of experiments utilized the most common ARM publicly available datasets to justify this hypothesis.

The UCI ML datasets, listed in Table~\ref{tab:exp_datasets}, were used for evaluating the performance of the proposed method~\cite{dua2019uci}. Each database is characterized by the number of transactions, number of attributes and their types, which can be either categorical (discrete) or numerical (real). These datasets were selected since they vary in terms of the number of transactions, the types of attributes, and the total number of attributes they contain. It is also worth mentioning that the proposed method determines the most suitable preprocessing algorithm automatically as part of its process, therefore, no manual preprocessing was applied to the original datasets.
\begin{table}[htb]
    \centering
    \caption{Evaluation datasets used in the experiments}
    \label{tab:exp_datasets}
    \begin{tabular}{lrrr}
    \hline
       Dataset  & Nr. of inst. & Nr. of attr. & Attr. type [D/N] \\
       \hline
       Abalone & 4,177 & 9 & DN \\
       Balance scale & 625 & 5 & DN \\
       Basketball & 96 & 5 & N \\
       Bolts& 40 & 8 & N \\
       Buying & 100 & 40 & N \\
       German & 1,000 & 20 & DN\\
       House & 22,784 & 17 & N\\       
       Ionosphere & 351 & 35 & DN \\
       Quake & 2,178 & 4 & N \\
       Wine & 178 & 14 & N \\
       \hline
    \end{tabular}
\end{table}

In our experiments, we used two outer commonly used Nia-s for the ARM pipeline optimization, namely, the DE and the PSO. To ensure a fair comparison, the most important hyper-parameters of both algorithms were set equally. The population size was set to $\mathit{NP}=30$, and the maximum number of fitness function evaluations to $\mathit{MAXFES}=1000$ (i.e., the number of pipeline evaluations). The other parameters were set to default values, as proposed in the Niapy library. In all the experiments, the inner optimization algorithms for mining association rules were selected similarly as in the example illustrated in Fig.~\ref{fig:decode_pipeline}. 
Each experimental run produced the best pipeline for a combination of the specific dataset and algorithm. Considering the stochastic nature of the DE and PSO algorithms, the reported results are the average fitness function values of the best obtained pipelines over 30 independent runs. 

The quality of the constructed pipeline was evaluated regarding Eq.~(\ref{eq:surrogate_fitness}) in the outer algorithm, while the fitness function in the inner algorithm was calculated as a weighted sum of the ARM metrics decoded from the corresponding individual by the NiaAutoARM.

\subsection{Results}

The following experiments were conducted for analyzing the newly proposed NiaAutoARM thoroughly:
\begin{itemize}
    \item baseline ARM pipeline optimization, allowing just one preprocessing component and disabling the ARM metric weight adaptation,
    \item influence of adapting the ARM metric weights on the quality of the ARM pipeline construction, 
    \item influence of selecting more preprocessing components on the quality of the ARM pipeline construction,
    \item comparison with the VARDE state-of-the-art algorithm.
\end{itemize}

In the remainder of this section, all the experimental results are presented in detail, showcasing the usefulness and efficiency of the proposed method.

\subsubsection{Baseline ARM pipeline construction}
The purpose of the first experiment was to establish a foundational comparison for all the subsequent experiments. In this experiment, no ARM metric weight adaptation was applied, ensuring that the generated pipelines operated in their default configurations. Additionally, each generated pipeline was restricted to a single preprocessing method, eliminating the variability introduced by multiple preprocessing components. 

\begin{figure*}[htb]
    \caption{Results for baseline ARM pipeline optimization, reporting the averages of best pipelines in terms of fitness values, number of generated rules, and the used inner optimization algorithms.}
    \label{fig:combined_sf}
    \centering
    \begin{subfigure}{0.45\textwidth}
    \centering
        \includegraphics[width=1\linewidth]{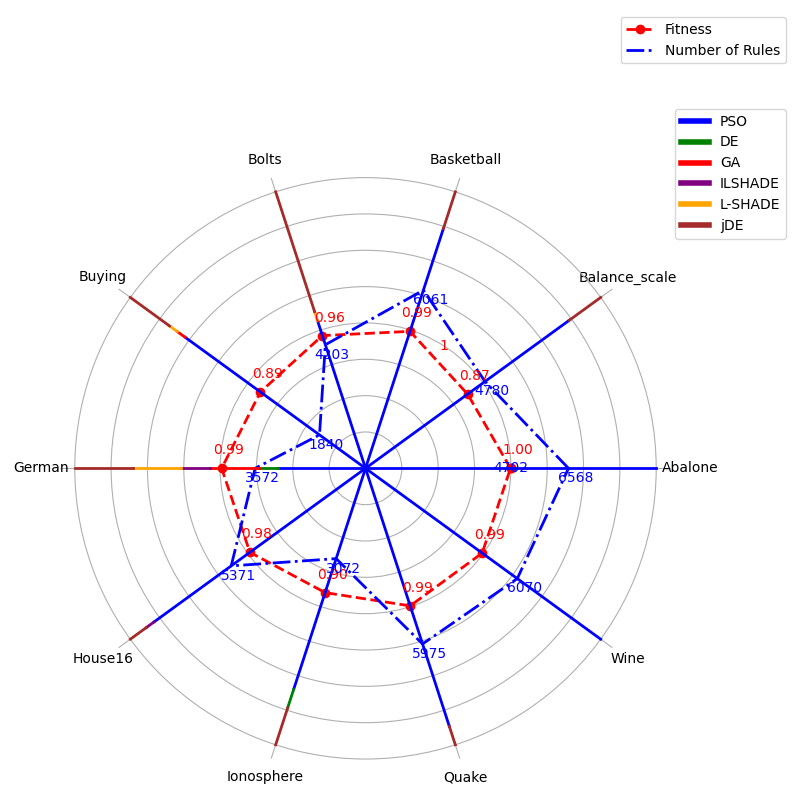}
    \caption{Results for outer algorithm PSO without ARM metric weight adaptation and just one preprocessing method.}
    \label{fig:res_one_preproc_pso}
    \end{subfigure}
    \hfill
    \begin{subfigure}{0.45\textwidth}
    \centering
    \includegraphics[width=1\linewidth]{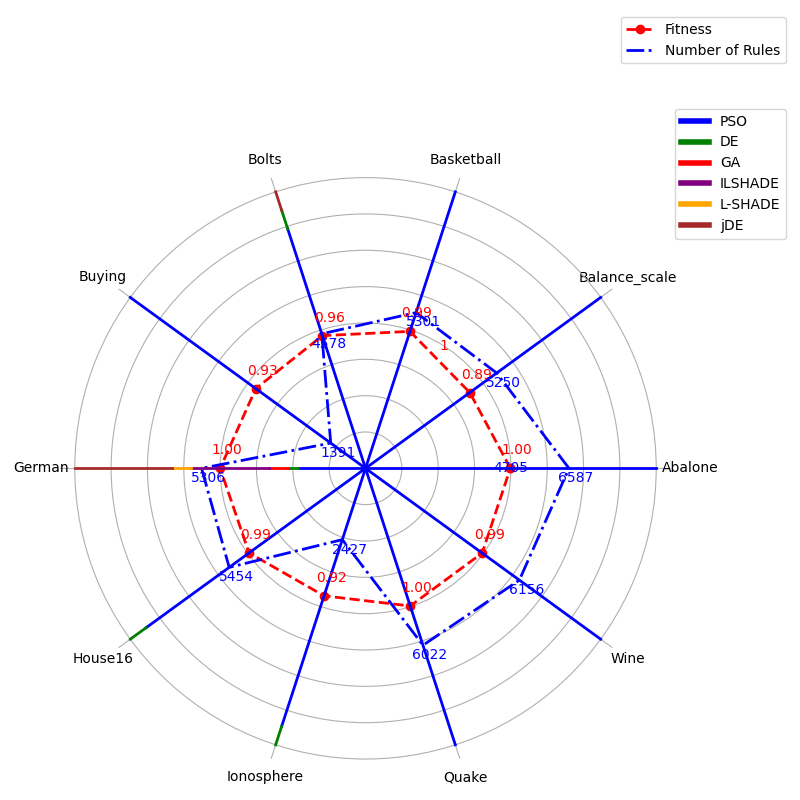}
    \caption{Results for outer algorithm DE without ARM metric weight adaptation and just one preprocessing method.}
    \label{fig:res_one_preproc_de}
    \end{subfigure}
\end{figure*}

All the results for this experiment are reported numerically in Tables~\ref{tab:res_one_preproc_PSO} and~\ref{tab:res_one_preproc_DE}, and graphically in Figure~\ref{fig:combined_sf} for the different PSO and DE outer meta-heuristics, respectively. The mentioned Tables are structured as follows: The column 'Preprocessing method' denotes the frequency of the preprocessing algorithms in the best obtained pipelines over all 30 runs. The column 'Hyper-parameters' is used for reporting the average obtained population sizes ($\mathit{NP}$) and maximum function evaluations ($\mathit{MAXFES}$) for the best obtained ARM pipelines. Lastly, the column 'Metrics \& Weights' are used for reporting the average values of each used ARM evaluation metric. The number in the subscript denotes the number of pipelines in which a specific metric was used. Since, in the baseline experiment, no ARM metric weight adaptation was used, all values are equal to 1. Each row in the Tables refer to one experimental dataset. 

Figure~\ref{fig:combined_sf} presents the obtained average fitness values, along with the average number of rules generated by the best obtained pipelines. Additionally, the frequencies of the inner optimization algorithms are depicted. The fitness values are marked with blue dash/dotted lines, whereas the number of rules is marked with a red dotted line. The frequencies of the inner algorithms are presented as different colored lines from the center of the graph, outward to each dataset. 

The results in Table~\ref{tab:res_one_preproc_PSO}, developed by the outer algorithm PSO, justified that the preprocessing methods, like  MM, ZS, and RHC, were selected more frequently, while, in general, 'No preprocessing' was selected in most of the pipelines regardless of the dataset. The ARM metrics support, confidence, and coverage appeared consistently across most datasets. Notably, the support and confidence are present in nearly all the pipelines for datasets, like Abalone, Balance scale, and Basketball, indicating that these metrics are essential for the underlying optimization process. Metrics like amplification, which were used less frequently, are absent in many datasets, suggesting that the current algorithm configuration does not prioritize such metrics. The hyper-parameters $\mathit{NP}$  and $\mathit{MAXFES}$ varied depending on the dataset, influencing the ARM pipeline optimization process.

Table~\ref{tab:res_one_preproc_DE} shows the results for the outer algorithm DE. Similar to the results of the PSO, key ARM metrics like support, confidence, and coverage are found consistently in many of the generated pipelines. However, there are subtle differences in the distribution of these metrics across the pipelines. For instance, the metric amplitude is selected just for the dataset German. Regarding the preprocessing methods and hyper-parameters, a similar distribution can be found as in the results of the PSO algorithm. 

The graphical results showcase that both DE and PSO obtained similar results regarding the fitness value. The number of rules is slightly dispersed, although no big deviations are detected. The key differences are in the selection of the inner optimization algorithm. For the majority of datasets, the PSO and jDE algorithms were selected more often as the inner optimization algorithms, than others. This is true for both the outer algorithm experiment runs. Other used algorithms, such as GA, DE, ILSHADE and LSHADE, were selected rarely, probably due to their complexity or their lack of it.

To summarize the results of the baseline experiment, we can conclude that the best results were obtained, when either no preprocessing was applied, or MM was used on the dataset. The $\mathit{NP}$ parameter seems to be higher for more complex datasets (i.e., more attributes) such as Buying, German, House16 and Ionosphere, while it remains lower for the others which were less demanding. Regarding the selection of specific ARM evaluation metrics, it seems that both algorithms focused on the more common ones, usually used in Evolutionary ARM~\cite{mlakar2023variable}.
Overall, these results indicate the DE and PSO algorithms' robustness as an outer ARM meta-heuristic, while reinforcing the potential benefits of further exploration into ARM metric weight adaptation and diversified preprocessing strategies. 

Let us notice that all the subsequent results are reported in the same manner.

\begin{table*}[htb]
\caption{Results for PSO algorithm, with $P=1$, without ARM metric weight adaptation.}
\label{tab:res_one_preproc_PSO}
\centering
\resizebox{\textwidth}{!}{
\begin{tabular}{c|cccccc|c|c|cccccc}
\hline
\multirow{2}{*}{Dataset} & \multicolumn{6}{c|}{Preprocessing method} & \multicolumn{2}{c|}{Hyper-parameters} & \multicolumn{6}{c|}{Metrics \& Weights}\\ \cline{2-15} 
& $MM$ & $ZS$ & $DS$ & $RHC$ & $KM$ & $N^{a}$ & $\mathit{NP}$ & $\mathit{MAXFES}$ & Supp & Conf & Cover & Amp & Incl & Comp\\ \hline
Abalone & 0.27 & 0.07 & - & 0.20 & - & 0.47 & 11.7 $\pm$ 5.2 & 9656.2 $\pm$ 796.4 & 1.00 $\pm 0.00_{25}$ & 1.00 $\pm 0.00_{23}$ & 1.00 $\pm 0.00_{19}$ & - & 1.00 $\pm 0.00_{22}$ & 1.00 $\pm 0.00_{15}$ \\
Balance scale & 0.30 & 0.07 & - & 0.10 & - & 0.53 & 17.6 $\pm$ 9.0 & 8370.3 $\pm$ 2598.6 & 1.00 $\pm 0.00_{24}$ & 1.00 $\pm 0.00_{28}$ & 1.00 $\pm 0.00_{8}$ & - & 1.00 $\pm 0.00_{19}$ & 1.00 $\pm 0.00_{16}$ \\
Basketball & 0.47 & - & - & - & - & 0.53 & 11.7 $\pm$ 4.8 & 9851.2 $\pm$ 543.7 & 1.00 $\pm 0.00_{25}$ & 1.00 $\pm 0.00_{26}$ & 1.00 $\pm 0.00_{18}$ & - & 1.00 $\pm 0.00_{29}$ & 1.00 $\pm 0.00_{12}$ \\
Bolts & 0.23 & 0.10 & - & 0.07 & - & 0.60 & 13.2 $\pm$ 6.6 & 8946.9 $\pm$ 2189.4 & 1.00 $\pm 0.00_{23}$ & 1.00 $\pm 0.00_{25}$ & 1.00 $\pm 0.00_{16}$ & 1.00 $\pm 0.00_{2}$ & 1.00 $\pm 0.00_{26}$ & 1.00 $\pm 0.00_{8}$ \\
Buying & 0.23 & 0.20 & - & - & 0.07 & 0.50 & 17.5 $\pm$ 8.3 & 9039.1 $\pm$ 1742.6 & 1.00 $\pm 0.00_{25}$ & 1.00 $\pm 0.00_{18}$ & 1.00 $\pm 0.00_{11}$ & 1.00 $\pm 0.00_{3}$ & 1.00 $\pm 0.00_{2}$ & 1.00 $\pm 0.00_{8}$ \\
German & - & - & 0.97 & - & - & 0.03 & 20.1 $\pm$ 7.2 & 5871.8 $\pm$ 3046.2 & 1.00 $\pm 0.00_{16}$ & 1.00 $\pm 0.00_{20}$ & 1.00 $\pm 0.00_{12}$ & 1.00 $\pm 0.00_{12}$ & 1.00 $\pm 0.00_{15}$ & 1.00 $\pm 0.00_{17}$ \\
House16 & 0.30 & 0.13 & - & 0.10 & - & 0.47 & 15.5 $\pm$ 8.3 & 8642.5 $\pm$ 2038.8 & 1.00 $\pm 0.00_{25}$ & 1.00 $\pm 0.00_{21}$ & 1.00 $\pm 0.00_{22}$ & 1.00 $\pm 0.00_{3}$ & 1.00 $\pm 0.00_{7}$ & 1.00 $\pm 0.00_{12}$ \\
Ionosphere & 0.17 & - & - & 0.03 & 0.10 & 0.70 & 18.3 $\pm$ 9.2 & 8600.9 $\pm$ 2393.7 & 1.00 $\pm 0.00_{28}$ & 1.00 $\pm 0.00_{19}$ & 1.00 $\pm 0.00_{7}$ & 1.00 $\pm 0.00_{1}$ & 1.00 $\pm 0.00_{3}$ & 1.00 $\pm 0.00_{2}$ \\
Quake & 0.30 & 0.03 & - & 0.07 & - & 0.60 & 11.4 $\pm$ 4.3 & 9622.0 $\pm$ 1074.7 & 1.00 $\pm 0.00_{27}$ & 1.00 $\pm 0.00_{24}$ & 1.00 $\pm 0.00_{17}$ & 1.00 $\pm 0.00_{1}$ & 1.00 $\pm 0.00_{17}$ & 1.00 $\pm 0.00_{18}$ \\
Wine & 0.23 & 0.03 & - & 0.10 & - & 0.63 & 12.6 $\pm$ 6.3 & 9471.0 $\pm$ 1301.1 & 1.00 $\pm 0.00_{24}$ & 1.00 $\pm 0.00_{25}$ & 1.00 $\pm 0.00_{20}$ & 1.00 $\pm 0.00_{1}$ & 1.00 $\pm 0.00_{14}$ & 1.00 $\pm 0.00_{18}$ \\
\cline{8-15}
&&&&&&&$14.94 \pm 3.06$& $8807.19\pm 1089.31$ &1.00 $\pm 0.00_{24.20\pm3.06}$& 1.00 $\pm 0.00_{22.90\pm3.11}$&1.00 $\pm 0.00_{15.00\pm4.92}$&1.00 $\pm 0.00_{3.29\pm3.65}$&1.00 $\pm 0.00_{15.40\pm8.73}$&1.00 $\pm 0.00_{12.60\pm5.00}$\\
\hline
\end{tabular}}
\footnotesize{$^a$ No preprocessing of the dataset}\\
\end{table*}

\begin{table*}[htb]
\caption{Results for DE algorithm, with $P=1$, without ARM metrics weight adaptation.}
\label{tab:res_one_preproc_DE}
\centering
\resizebox{\textwidth}{!}{
\begin{tabular}{c|cccccc|c|c|cccccc}
\hline
\multirow{2}{*}{Dataset} & \multicolumn{6}{c|}{Preprocessing method} & \multicolumn{2}{c|}{Hyper-parameters} & \multicolumn{6}{c|}{Metrics \& Weights}\\ \cline{2-15} 
& $MM$ & $ZS$ & $DS$ & $RHC$ & $KM$ & $N^{a}$ & $\mathit{NP}$ & $\mathit{MAXFES}$ & Supp & Conf & Cover & Amp & Incl & Comp\\ \hline
Abalone & 0.43 & 0.10 & - & 0.20 & - & 0.27 & 13.2 $\pm$ 5.4 & 9360.9 $\pm$ 1150.8 & 1.00 $\pm 0.00_{27}$ & 1.00 $\pm 0.00_{22}$ & 1.00 $\pm 0.00_{20}$ & - & 1.00 $\pm 0.00_{23}$ & 1.00 $\pm 0.00_{8}$ \\
Balance scale & 0.33 & 0.07 & - & 0.20 & - & 0.40 & 14.8 $\pm$ 7.4 & 8216.3 $\pm$ 2234.2 & 1.00 $\pm 0.00_{23}$ & 1.00 $\pm 0.00_{28}$ & 1.00 $\pm 0.00_{7}$ & - & 1.00 $\pm 0.00_{23}$ & 1.00 $\pm 0.00_{20}$ \\
Basketball & 0.47 & 0.17 & - & 0.13 & - & 0.23 & 12.9 $\pm$ 3.7 & 9160.8 $\pm$ 1468.8 & 1.00 $\pm 0.00_{22}$ & 1.00 $\pm 0.00_{25}$ & 1.00 $\pm 0.00_{17}$ & - & 1.00 $\pm 0.00_{28}$ & 1.00 $\pm 0.00_{9}$ \\
Bolts & 0.27 & 0.13 & - & 0.10 & - & 0.50 & 15.4 $\pm$ 6.3 & 9107.0 $\pm$ 1343.4 & 1.00 $\pm 0.00_{25}$ & 1.00 $\pm 0.00_{21}$ & 1.00 $\pm 0.00_{15}$ & - & 1.00 $\pm 0.00_{23}$ & 1.00 $\pm 0.00_{10}$ \\
Buying & 0.33 & 0.17 & - & 0.10 & 0.10 & 0.30 & 13.8 $\pm$ 6.6 & 8793.3 $\pm$ 1813.5 & 1.00 $\pm 0.00_{28}$ & 1.00 $\pm 0.00_{13}$ & 1.00 $\pm 0.00_{6}$ & - & 1.00 $\pm 0.00_{1}$ & - \\
German & - & - & 1.00 & - & - & - & 18.7 $\pm$ 7.4 & 7992.1 $\pm$ 2403.1 & 1.00 $\pm 0.00_{16}$ & 1.00 $\pm 0.00_{13}$ & 1.00 $\pm 0.00_{15}$ & 1.00 $\pm 0.00_{19}$ & 1.00 $\pm 0.00_{15}$ & 1.00 $\pm 0.00_{13}$ \\
House16 & 0.50 & 0.20 & - & 0.03 & - & 0.27 & 14.2 $\pm$ 6.4 & 8751.8 $\pm$ 1865.9 & 1.00 $\pm 0.00_{23}$ & 1.00 $\pm 0.00_{25}$ & 1.00 $\pm 0.00_{23}$ & - & 1.00 $\pm 0.00_{8}$ & 1.00 $\pm 0.00_{17}$ \\
Ionosphere & 0.30 & - & - & 0.10 & 0.10 & 0.50 & 15.1 $\pm$ 6.6 & 8769.9 $\pm$ 2080.5 & 1.00 $\pm 0.00_{30}$ & 1.00 $\pm 0.00_{19}$ & 1.00 $\pm 0.00_{4}$ & - & - & 1.00 $\pm 0.00_{2}$ \\
Quake & 0.13 & 0.23 & - & 0.17 & - & 0.47 & 11.1 $\pm$ 2.9 & 9406.5 $\pm$ 899.3 & 1.00 $\pm 0.00_{24}$ & 1.00 $\pm 0.00_{18}$ & 1.00 $\pm 0.00_{18}$ & - & 1.00 $\pm 0.00_{21}$ & 1.00 $\pm 0.00_{15}$ \\
Wine & 0.27 & 0.07 & - & 0.13 & - & 0.53 & 11.8 $\pm$ 2.8 & 9506.5 $\pm$ 827.2 & 1.00 $\pm 0.00_{27}$ & 1.00 $\pm 0.00_{28}$ & 1.00 $\pm 0.00_{15}$ & - & 1.00 $\pm 0.00_{21}$ & 1.00 $\pm 0.00_{10}$ \\
\cline{8-15}
&&&&&&&$14.09 \pm 2.03$& $8906\pm478.47$&1.00 $\pm 0.00_{24.50\pm3.72}$& 1.00 $\pm 0.00_{21.20\pm5.21}$&1.00 $\pm 0.00_{14.00\pm5.98}$&1.00 $\pm 0.00_{19.00\pm0.00}$&1.00 $\pm 0.00_{18.11\pm8.10}$&1.00 $\pm 0.00_{11.56\pm5.06}$\\
\hline
\end{tabular}}
\end{table*}

\subsubsection{Influence of the ARM metric weights adaption on the quality of ARM pipeline construction}
The purpose of this experiment was to analyze the impact of selecting ARM metric weight adaptation on the performance of the ARM pipeline construction. The ARM metric weights play a crucial role in guiding the optimization process, as they influence the evaluation and selection of the candidate association rules. By incorporating the ARM weight adaptation mechanism, the pipeline can adjust the importance of ARM metrics dynamically, such as support, confidence, coverage, and others, tailored to the characteristics of the dataset. This experiment aimed to determine whether adapting these weights improved the quality of the discovered rules, which are, therefore, reflected in the pipeline's metrics. The results are compared to the baseline configuration, where no weight adaptation was applied.

Tables~\ref{tab:res_one_preproc_ow_PSO} and~\ref{tab:res_one_preproc_ow_DE} present the results obtained by the outer algorithms PSO and DE, respectively. The similar selection of the preprocessing methods as in the last experiment was also employed in this experiment, where the preprocessing methods MM, ZS and None were applied the most frequently. The hyper-parameters yielded higher values for the harder datasets. Considering the ARM metrics, the support and confidence still arose with high weight values in the majority of the pipelines, whereas the ARM metrics, like amplification or comprehensibility, are utilized less with lower weights. 

\begin{figure*}[htb]
    \caption{Results of ARM pipeline construction using ARM metric weight adaptation, reporting the averages of best pipelines in terms of fitness values, number of generated rules, and the used inner optimization algorithms.}
    \label{fig:combined_sf_ow}
    \centering
    \begin{subfigure}{0.45\textwidth}
    \centering
    \includegraphics[width=1\linewidth]{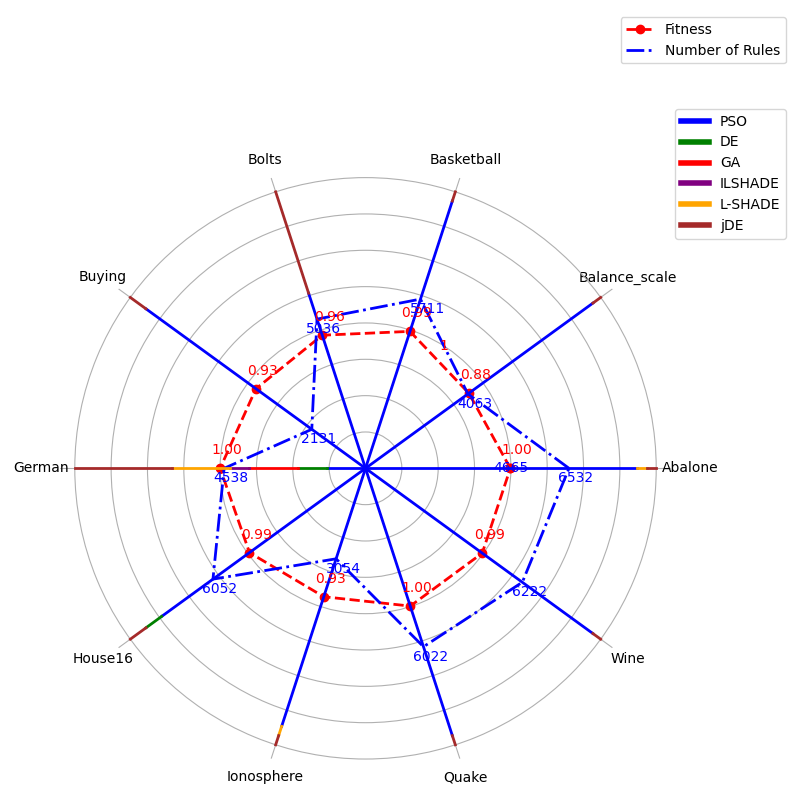}
    \caption{Results for outer algorithm PSO with ARM metric weight adaptation and just one preprocessing method.}
    \label{fig:res_one_preproc_ow_PSO}
    \end{subfigure}
    \hfill
    \begin{subfigure}{0.45\textwidth}
    \centering
    \includegraphics[width=1\linewidth]{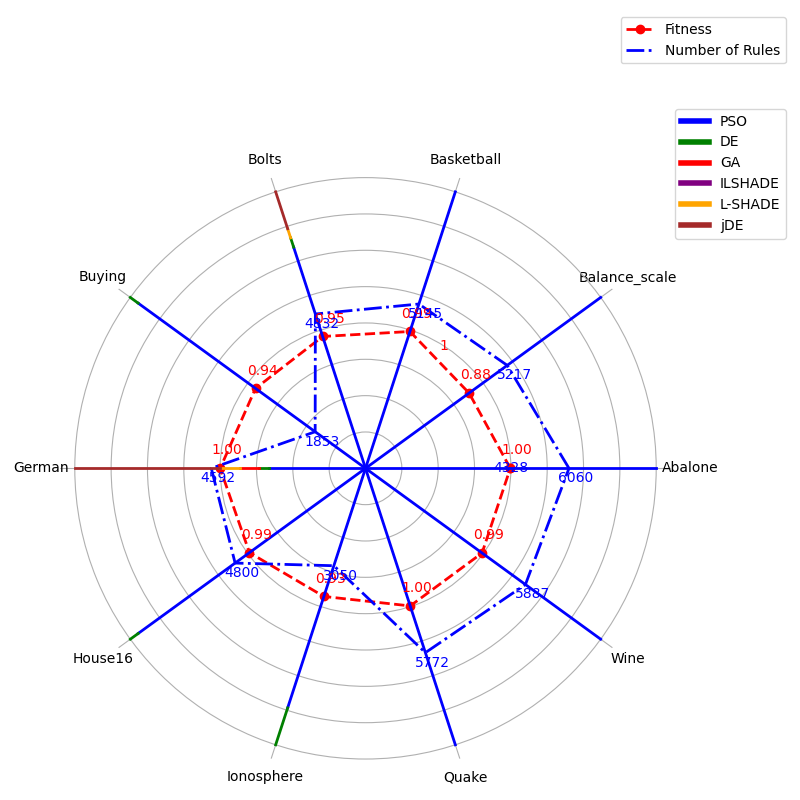}
    \caption{Results for outer algorithm DE with ARM metric weight adaptation and just one preprocessing method.}
    \label{fig:res_one_preproc_ow_DE}
    \end{subfigure}
\end{figure*}

From the results in Figure~\ref{fig:combined_sf_ow} we can deduce similar conclusions as from those in the baseline experiment, but the ARM metric weight adaptation provided slightly higher fitness values then those achieved in the last experiment. Although these differences are not significantly different according to the Wilcox test (p-value=0.41), they still offer overall better ARM pipelines for the majority of datasets.

\begin{table*}[htb]
\caption{Results for outer algorithm PSO with ARM metric weight adaptation and just one preprocessing method.}
\label{tab:res_one_preproc_ow_PSO}
\centering
\resizebox{\textwidth}{!}{
\begin{tabular}{c|cccccc|c|c|cccccc}
\hline
\multirow{2}{*}{Dataset} & \multicolumn{6}{c|}{Preprocessing method} & \multicolumn{2}{c|}{Hyper-parameters} & \multicolumn{6}{c|}{Metrics \& Weights}\\ \cline{2-15} 
& $MM$ & $ZS$ & $DS$ & $RHC$ & $KM$ & $N^{a}$ & $\mathit{NP}$ & $\mathit{MAXFES}$ & Supp & Conf & Cover & Amp & Incl & Comp\\ \hline
Abalone & 0.40 & 0.07 & - & 0.10 & - & 0.43 & 11.6 $\pm$ 5.0 & 9448.7 $\pm$ 1608.1 & 0.89 $\pm 0.23_{23}$ & 0.81 $\pm 0.29_{25}$ & 0.67 $\pm 0.33_{17}$ & - & 0.63 $\pm 0.35_{23}$ & 0.41 $\pm 0.29_{11}$ \\
Balance scale & 0.40 & 0.10 & - & 0.10 & - & 0.40 & 16.6 $\pm$ 8.8 & 6563.6 $\pm$ 3507.9 & 0.56 $\pm 0.39_{23}$ & 0.77 $\pm 0.30_{23}$ & 0.62 $\pm 0.25_{8}$ & - & 0.66 $\pm 0.31_{14}$ & 0.74 $\pm 0.27_{15}$ \\
Basketball & 0.63 & - & - & 0.07 & - & 0.30 & 14.8 $\pm$ 7.9 & 9285.8 $\pm$ 1723.5 & 0.83 $\pm 0.28_{29}$ & 0.84 $\pm 0.22_{24}$ & 0.63 $\pm 0.36_{10}$ & - & 0.76 $\pm 0.34_{22}$ & 0.88 $\pm 0.24_{9}$ \\
Bolts & 0.23 & 0.07 & - & 0.03 & - & 0.67 & 10.9 $\pm$ 3.6 & 8642.9 $\pm$ 2285.0 & 0.86 $\pm 0.21_{19}$ & 0.68 $\pm 0.32_{19}$ & 0.75 $\pm 0.28_{15}$ & - & 0.84 $\pm 0.27_{25}$ & 0.98 $\pm 0.04_{5}$ \\
Buying & 0.43 & 0.03 & - & 0.13 & 0.03 & 0.37 & 17.6 $\pm$ 8.4 & 8695.0 $\pm$ 2184.4 & 0.75 $\pm 0.31_{27}$ & 0.83 $\pm 0.33_{13}$ & 0.61 $\pm 0.40_{6}$ & 1.00 $\pm 0.00_{1}$ & 0.98 $\pm 0.00_{1}$ & 0.99 $\pm 0.01_{2}$ \\
German & - & - & 1.00 & - & - & - & 20.4 $\pm$ 7.0 & 5921.3 $\pm$ 2437.6 & 0.53 $\pm 0.28_{13}$ & 0.60 $\pm 0.35_{14}$ & 0.47 $\pm 0.36_{15}$ & 0.62 $\pm 0.36_{11}$ & 0.66 $\pm 0.35_{15}$ & 0.61 $\pm 0.29_{19}$ \\
House16 & 0.30 & 0.03 & - & 0.03 & - & 0.63 & 13.7 $\pm$ 6.5 & 9141.0 $\pm$ 1947.1 & 0.79 $\pm 0.28_{24}$ & 0.88 $\pm 0.20_{18}$ & 0.62 $\pm 0.36_{14}$ & 0.03 $\pm 0.03_{4}$ & 0.67 $\pm 0.46_{6}$ & 0.41 $\pm 0.29_{10}$ \\
Ionosphere & 0.23 & - & - & 0.13 & 0.03 & 0.60 & 13.3 $\pm$ 6.1 & 8799.0 $\pm$ 2451.1 & 0.77 $\pm 0.35_{29}$ & 0.68 $\pm 0.34_{20}$ & 0.73 $\pm 0.22_{3}$ & 0.03 $\pm 0.00_{1}$ & - & 0.34 $\pm 0.20_{2}$ \\
Quake & 0.40 & - & - & 0.13 & - & 0.47 & 12.1 $\pm$ 5.9 & 9941.2 $\pm$ 239.3 & 0.80 $\pm 0.29_{25}$ & 0.74 $\pm 0.34_{16}$ & 0.83 $\pm 0.21_{15}$ & - & 0.72 $\pm 0.32_{17}$ & 0.87 $\pm 0.29_{13}$ \\
Wine & 0.37 & 0.07 & - & 0.03 & - & 0.53 & 10.8 $\pm$ 2.0 & 9454.8 $\pm$ 1539.8 & 0.85 $\pm 0.24_{23}$ & 0.88 $\pm 0.26_{25}$ & 0.74 $\pm 0.30_{10}$ & - & 0.73 $\pm 0.33_{24}$ & 0.70 $\pm 0.23_{6}$ \\
\cline{8-15}
&&&&&&&$14.16 \pm 3.00$& $8589.34\pm1240.34$&0.76 $\pm 0.12_{23.50\pm4.54}$& 0.77 $\pm 0.09_{19.70\pm4.24}$&0.67 $\pm 0.10_{11.30\pm4.38}$&0.42 $\pm 0.41_{4.25\pm4.09}$&0.74 $\pm 0.10_{16.33\pm7.89}$&0.69 $\pm 0.23_{9.20\pm5.29}$\\
\hline
\end{tabular}}
\end{table*}

\begin{table*}[htb]
\caption{Results for outer algorithm DE with ARM metric weight adaptation and just one preprocessing method.}
\label{tab:res_one_preproc_ow_DE}
\centering
\resizebox{\textwidth}{!}{
\begin{tabular}{c|cccccc|c|c|cccccc}
\hline
\multirow{2}{*}{Dataset} & \multicolumn{6}{c|}{Preprocessing method} & \multicolumn{2}{c|}{Hyper-parameters} & \multicolumn{6}{c|}{Metrics \& Weights}\\ \cline{2-15} 
& $MM$ & $ZS$ & $DS$ & $RHC$ & $KM$ & $N^{a}$ & $\mathit{NP}$ & $\mathit{MAXFES}$ & Supp & Conf & Cover & Amp & Incl & Comp\\ \hline
Abalone & 0.60 & 0.03 & - & 0.10 & - & 0.27 & 12.1 $\pm$ 3.9 & 8808.1 $\pm$ 1628.4 & 0.78 $\pm 0.29_{24}$ & 0.84 $\pm 0.18_{20}$ & 0.67 $\pm 0.32_{19}$ & - & 0.63 $\pm 0.33_{26}$ & 0.65 $\pm 0.30_{11}$ \\
Balance scale & 0.37 & 0.10 & - & 0.03 & - & 0.50 & 19.3 $\pm$ 8.0 & 8727.0 $\pm$ 1780.1 & 0.66 $\pm 0.30_{25}$ & 0.80 $\pm 0.19_{20}$ & 0.66 $\pm 0.29_{9}$ & - & 0.85 $\pm 0.26_{15}$ & 0.66 $\pm 0.36_{15}$ \\
Basketball & 0.37 & 0.17 & - & 0.27 & - & 0.20 & 13.0 $\pm$ 5.1 & 8858.1 $\pm$ 1383.7 & 0.70 $\pm 0.31_{22}$ & 0.85 $\pm 0.21_{22}$ & 0.63 $\pm 0.26_{11}$ & - & 0.69 $\pm 0.34_{27}$ & 0.56 $\pm 0.33_{9}$ \\
Bolts & 0.17 & 0.20 & - & 0.07 & - & 0.57 & 16.1 $\pm$ 7.7 & 8495.1 $\pm$ 2678.4 & 0.67 $\pm 0.28_{20}$ & 0.59 $\pm 0.34_{23}$ & 0.69 $\pm 0.36_{16}$ & 0.25 $\pm 0.00_{1}$ & 0.59 $\pm 0.27_{23}$ & 0.78 $\pm 0.30_{12}$ \\
Buying & 0.27 & 0.13 & - & 0.10 & 0.07 & 0.43 & 16.8 $\pm$ 6.9 & 9124.0 $\pm$ 1631.4 & 0.73 $\pm 0.33_{30}$ & 0.68 $\pm 0.33_{14}$ & 0.69 $\pm 0.34_{3}$ & - & 0.10 $\pm 0.00_{1}$ & 0.49 $\pm 0.44_{2}$ \\
German & - & - & 1.00 & - & - & - & 19.4 $\pm$ 7.6 & 5848.0 $\pm$ 3014.7 & 0.87 $\pm 0.17_{12}$ & 0.88 $\pm 0.17_{10}$ & 0.57 $\pm 0.30_{10}$ & 0.61 $\pm 0.37_{11}$ & 0.63 $\pm 0.27_{12}$ & 0.59 $\pm 0.28_{10}$ \\
House16 & 0.33 & 0.07 & - & 0.23 & 0.03 & 0.33 & 15.4 $\pm$ 6.7 & 8682.6 $\pm$ 1810.6 & 0.64 $\pm 0.30_{23}$ & 0.76 $\pm 0.31_{16}$ & 0.67 $\pm 0.32_{20}$ & - & 0.41 $\pm 0.37_{9}$ & 0.60 $\pm 0.33_{18}$ \\
Ionosphere & 0.40 & - & - & 0.07 & 0.13 & 0.40 & 14.7 $\pm$ 6.4 & 8727.3 $\pm$ 1754.6 & 0.72 $\pm 0.27_{28}$ & 0.73 $\pm 0.32_{14}$ & 0.79 $\pm 0.23_{6}$ & - & 0.46 $\pm 0.40_{3}$ & 0.48 $\pm 0.17_{3}$ \\
Quake & 0.33 & 0.10 & - & 0.17 & - & 0.40 & 11.1 $\pm$ 2.6 & 9471.8 $\pm$ 1115.7 & 0.66 $\pm 0.32_{26}$ & 0.68 $\pm 0.25_{18}$ & 0.69 $\pm 0.30_{18}$ & - & 0.74 $\pm 0.30_{13}$ & 0.59 $\pm 0.33_{17}$ \\
Wine & 0.40 & 0.10 & - & 0.10 & - & 0.40 & 11.8 $\pm$ 4.0 & 9293.9 $\pm$ 1261.7 & 0.75 $\pm 0.26_{24}$ & 0.77 $\pm 0.24_{19}$ & 0.54 $\pm 0.35_{9}$ & - & 0.61 $\pm 0.34_{20}$ & 0.55 $\pm 0.34_{11}$ \\
\cline{8-15}
&&&&&&&$14.95 \pm 2.84$& $8603\pm961.74$&0.72 $\pm 0.07_{23.40\pm4.67}$& 0.76 $\pm 0.09_{17.60\pm3.85}$&0.66 $\pm 0.07_{12.10\pm5.52}$&0.43 $\pm 0.18_{6.00\pm5.00}$&0.57 $\pm 0.20_{14.90\pm8.62}$&0.60 $\pm 0.08_{10.80\pm5.02}$\\
\hline
\end{tabular}}
\end{table*}

\subsubsection{Influence of selecting more preprocessing methods on the quality of ARM pipeline construction}

The parameter $P$ controls the number of preprocessing components allowed in an ARM pipeline. By increasing $P$ beyond 1, we introduce the possibility of combining multiple preprocessing dataset methods, which can, potentially, enhance the quality of the generated rules. This increased flexibility enables the pipeline to address complex data characteristics (e.g., variability in feature scaling, noise reduction, or dimensionality reduction) more effectively. However, this increased complexity also poses challenges, including higher computational costs and a broader search space to be discovered by the inner optimization algorithms. In this section, we analyze the impact of setting the parameter as $P > 1$ on the quality of the ARM pipelines, focusing on the resulting ARM metrics and their corresponding weights, and computational trade offs for the experimental datasets. The results of the selected preprocessing algorithms are depicted as heatmaps of all the possible combinations. The results in Tables~\ref{tab:res_multi_preproc_ow_PSO} and~\ref{tab:res_multi_preproc_ow_DE} suggest that the support and confidence ARM metrics were again included heavily in the calculation of the fitness function, achieving high values in the majority of the pipelines for both the outer optimization algorithms. The coverage and inclusion ARM metrics were also involved in many pipelines, although their average weights were smaller. There was no notable difference in the selected hyper-parameters when compared to the previous two experiments. 

\begin{figure*}[htb]
    \caption{Results of PSO ARM pipeline optimization using ARM metric weight adaptation and selecting more preprocessing components, reporting the averages of best pipelines in terms of fitness values, number of generated rules, the used inner optimization algorithms and preprocessing methods.}
    \label{fig:combined_pso_sf_ow_amp}
    \centering
    \begin{subfigure}{0.45\textwidth}
    \centering
    \includegraphics[width=1\linewidth]{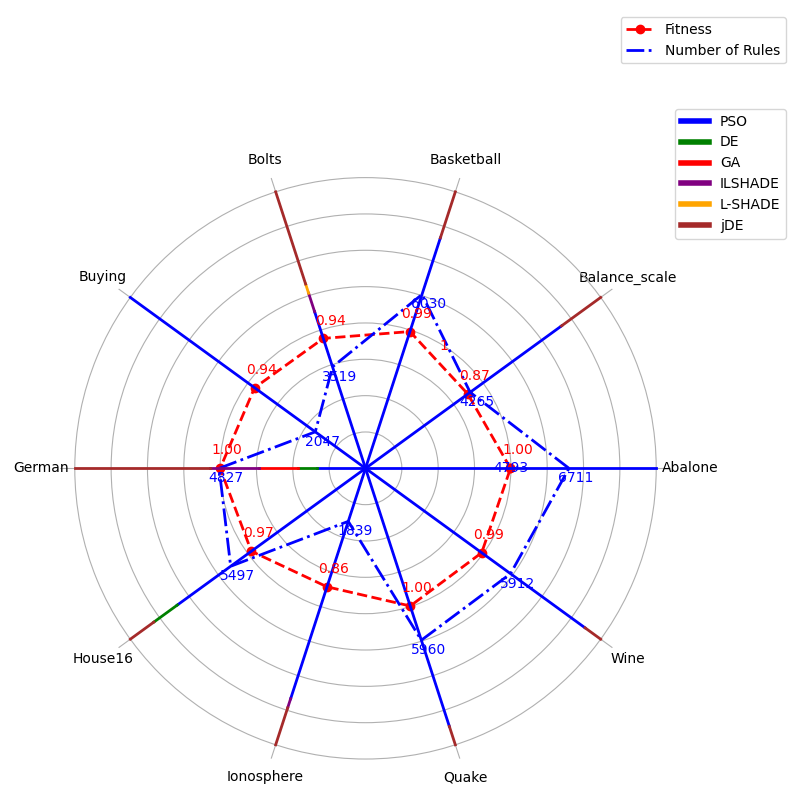}
    \caption{Results of preprocessing components for outer algorithm PSO with ARM metric weight adaptation and more preprocessing methods.}
    \label{fig:res_multi_preproc_ow_PSO}
    \end{subfigure}
    \hfill
    \begin{subfigure}{0.5\textwidth}
        \centering
    \includegraphics[width=1\linewidth]{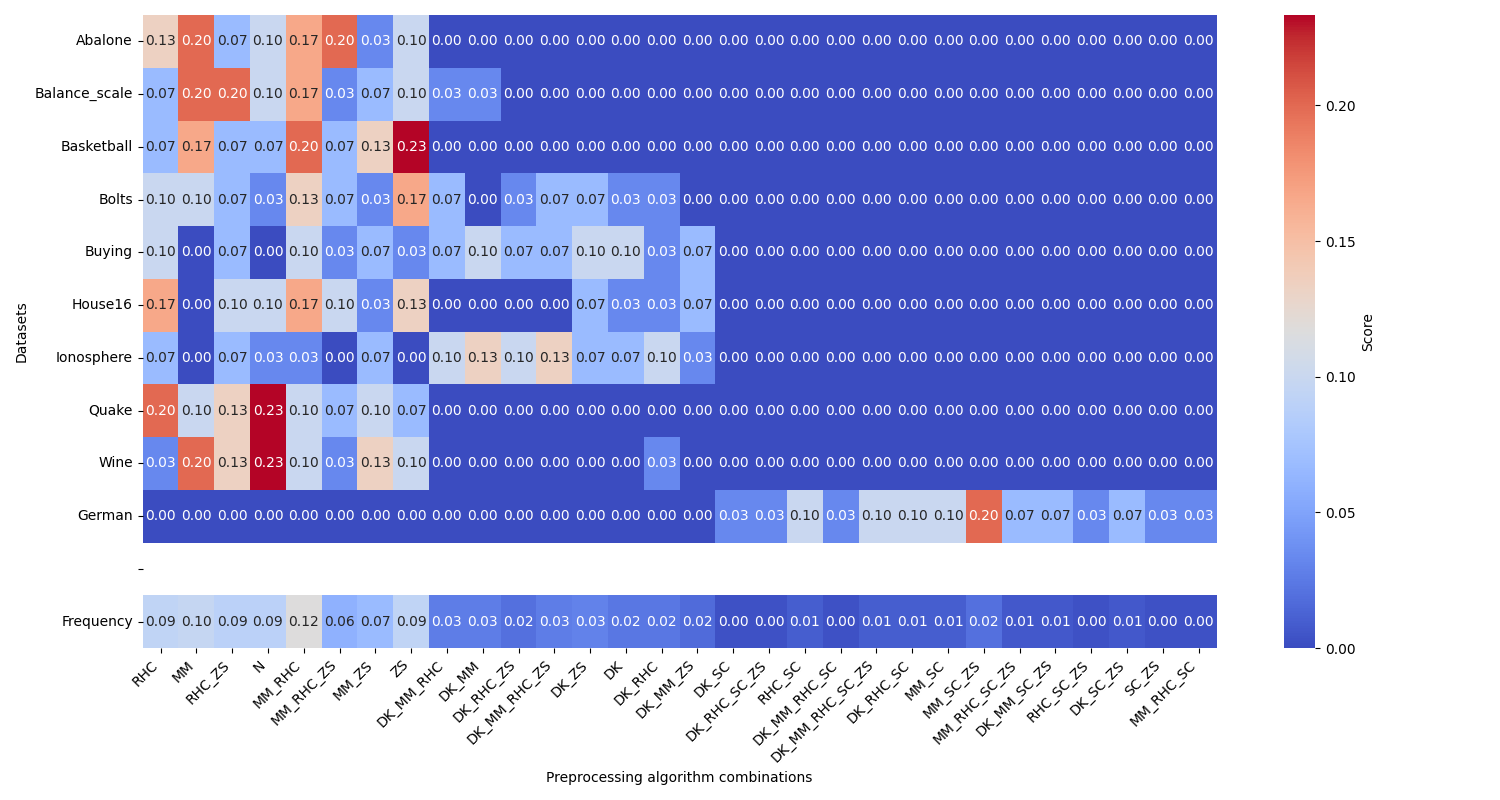}
    \caption{Heatmap of preprocessing components for outer algorithm PSO with ARM metric weight adaptation and more preprocessing methods.}
        \label{fig:heatmap_multi_preproc_ow_PSO}
    \end{subfigure}
\end{figure*}

\begin{figure*}[htb]
    \caption{Results of DE ARM pipeline optimization using ARM metric weight adaptation and selecting more preprocessing methods, reporting the averages of best pipelines in terms of fitness values, number of generated rules, the used inner optimization algorithms and preprocessing methods.}
    \label{fig:combined_de_sf_ow_amp}
    \centering
    \begin{subfigure}{0.45\textwidth}
        \centering
    \includegraphics[width=1\linewidth]{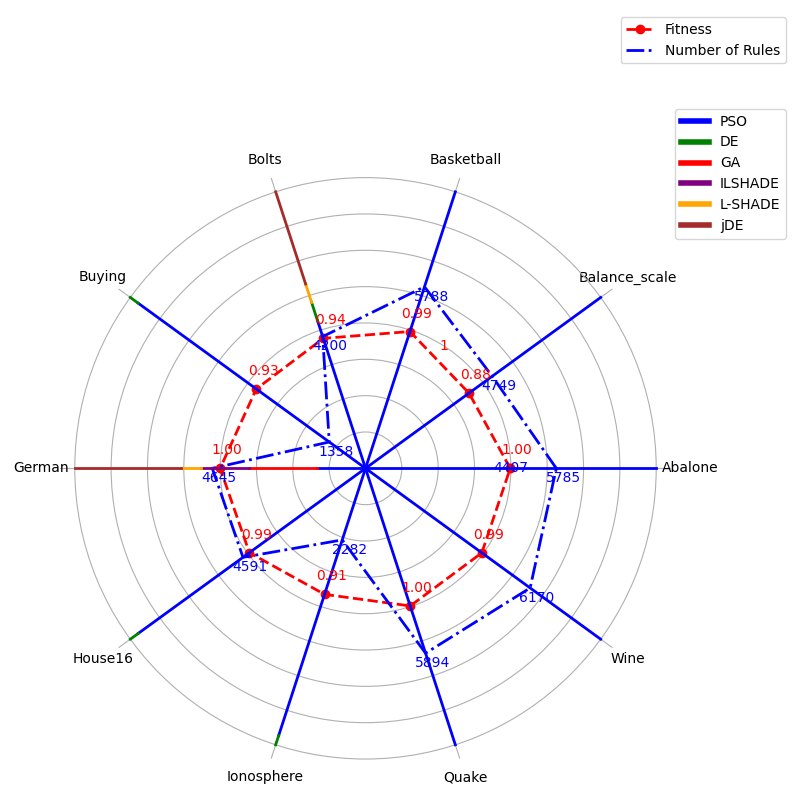}
    \caption{Results of preprocessing components for outer algorithm DE with ARM metric weight adaptation and more preprocessing methods.}
        \label{fig:res_multi_preproc_ow_DE}
    \end{subfigure}
    \hfill
    \begin{subfigure}{0.5\textwidth}
        \centering
    \includegraphics[width=1\linewidth]{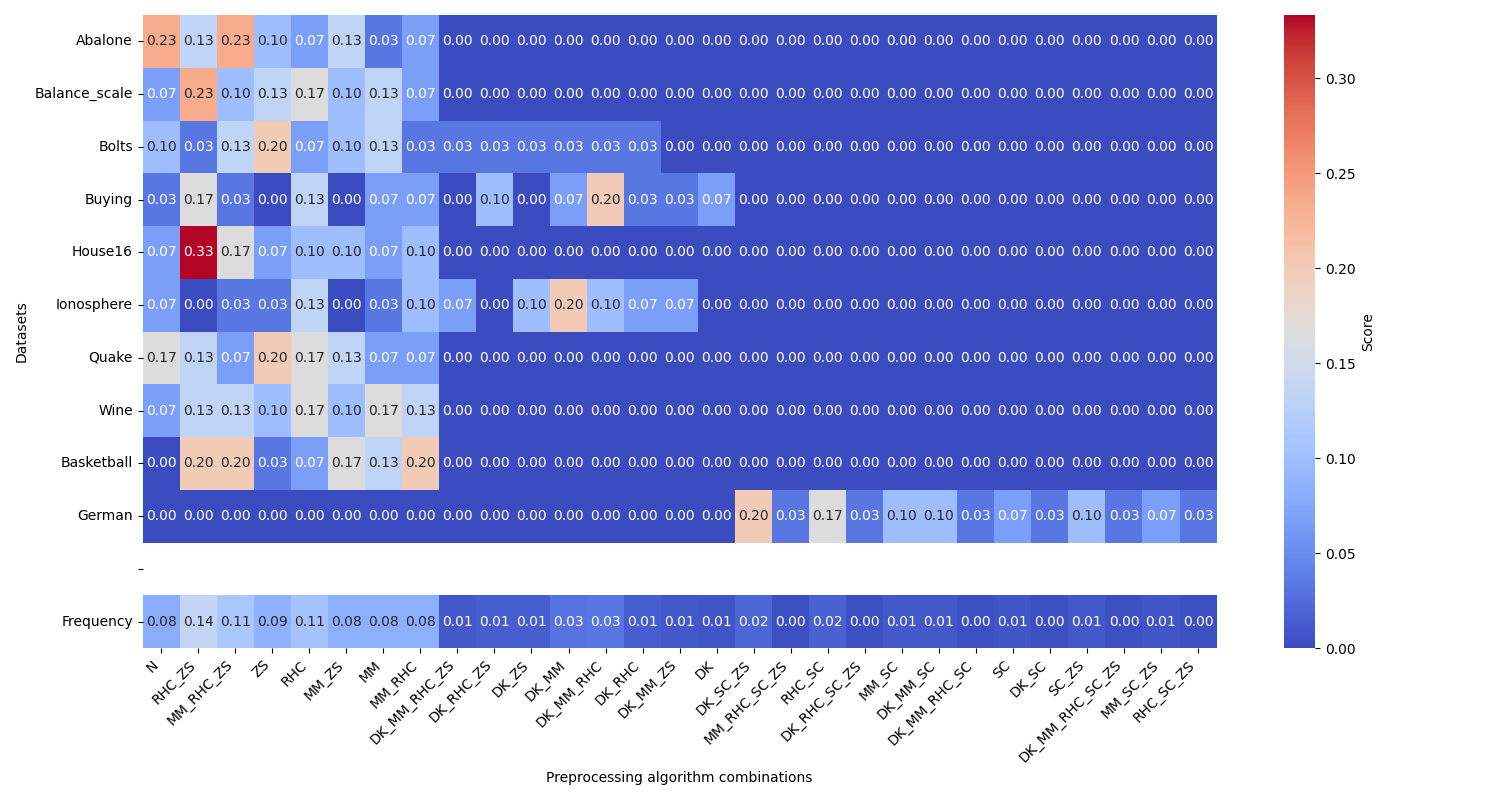}
    \caption{Heatmap of preprocessing components for outer algorithm DE with ARM metric weight adaptation and more preprocessing methods.}
        \label{fig:heatmap_multi_preproc_ow_DE}
    \end{subfigure}
\end{figure*}

Since this experiment included selecting more preprocessing methods, their selection frequency is reported in terms of heatmaps in Figure~\ref{fig:combined_pso_sf_ow_amp}b for the PSO algorithm, and Figure~\ref{fig:combined_de_sf_ow_amp}b for the DE algorithm, respectively. The selection of the preprocessing method varies, of course, if we observe a particular dataset, as the data are distributed differently. However, if we look at the overall selection process, specific combinations stand out. For the PSO algorithm the most frequent combinations are $\{\text{MM}, \text{RHC}\}$ and  MM, while, for the DE algorithm, the $\{\text{RHC}, \text{ZS}\}$, $\{\text{MM}, \text{RHC}, \text{ZS}\}$, and RHC. The MM preprocessing method was selected frequently across all datasets in both algorithms, likely due to its ability to normalize feature values to a standard range, which enhances the ability of the inner optimization algorithm to explore the search space more efficiently. This preprocessing method ensures that all features contribute equally during the optimization process, mitigating the influence of features with larger numeric ranges and facilitating better rule generation. 

Figures~\ref{fig:combined_pso_sf_ow_amp}a and~\ref{fig:combined_de_sf_ow_amp}a illustrate the fitness values and the number of generated rules for the PSO and DE algorithms. The DE algorithm produces ARM pipelines with slightly higher fitness values, while the PSO algorithm generates a greater number of rules. It is also evident that the PSO algorithm was selected the most as the inner optimization algorithm in both scenarios.

\begin{table*}[htb]
\caption{Results for outer algorithm PSO with ARM metric weight adaptation and selecting more preprocessing methods.}
\label{tab:res_multi_preproc_ow_PSO}
\centering
\resizebox{\textwidth}{!}{
\begin{tabular}{c|cccccc|c|c|cccccc}
\hline
\multirow{2}{*}{Dataset} & \multicolumn{6}{c|}{Preprocessing method} & \multicolumn{2}{c|}{Hyper-parameters} & \multicolumn{6}{c|}{Metrics \& Weights}\\ \cline{2-15} 
& $MM$ & $ZS$ & $DS$ & $RHC$ & $KM$ & $N^{a}$ & $\mathit{NP}$ & $\mathit{MAXFES}$ & Supp & Conf & Cover & Amp & Incl & Comp\\ \hline
Abalone & - & - & - & - & - & - & 15.6 $\pm$ 8.4 & 9570.9 $\pm$ 1477.9 & 0.83 $\pm 0.30_{17}$ & 0.82 $\pm 0.24_{21}$ & 0.83 $\pm 0.28_{19}$ & - & 0.65 $\pm 0.40_{17}$ & 0.76 $\pm 0.36_{16}$ \\
Balance scale & - & - & - & - & - & - & 14.8 $\pm$ 7.6 & 7869.1 $\pm$ 2986.2 & 0.69 $\pm 0.37_{23}$ & 0.74 $\pm 0.29_{24}$ & 0.48 $\pm 0.35_{10}$ & - & 0.82 $\pm 0.27_{16}$ & 0.70 $\pm 0.28_{14}$ \\
Basketball & - & - & - & - & - & - & 13.4 $\pm$ 6.5 & 9700.8 $\pm$ 907.4 & 0.73 $\pm 0.34_{24}$ & 0.83 $\pm 0.30_{19}$ & 0.88 $\pm 0.25_{11}$ & - & 0.66 $\pm 0.38_{21}$ & 0.76 $\pm 0.33_{10}$ \\
Bolts  & - & - & - & - & - & - & 15.7 $\pm$ 7.9 & 8379.6 $\pm$ 2697.1 & 0.79 $\pm 0.29_{25}$ & 0.86 $\pm 0.24_{18}$ & 0.82 $\pm 0.24_{14}$ & - & 0.76 $\pm 0.28_{21}$ & 0.79 $\pm 0.27_{8}$ \\
Buying & - & - & - & - & - & -  & 19.3 $\pm$ 8.7 & 9364.9 $\pm$ 1770.2 & 0.80 $\pm 0.29_{26}$ & 0.88 $\pm 0.21_{13}$ & 0.79 $\pm 0.32_{7}$ & - & - & 0.66 $\pm 0.23_{2}$ \\
German & - & - & - & - & - & - & 19.4 $\pm$ 6.5 & 6091.4 $\pm$ 3015.5 & 0.61 $\pm 0.29_{13}$ & 0.67 $\pm 0.30_{14}$ & 0.51 $\pm 0.38_{13}$ & 0.76 $\pm 0.28_{14}$ & 0.66 $\pm 0.31_{18}$ & 0.54 $\pm 0.33_{14}$ \\
House16 & - & - & - & - & - & - & 16.0 $\pm$ 8.3 & 8451.8 $\pm$ 2975.0 & 0.71 $\pm 0.33_{24}$ & 0.80 $\pm 0.29_{22}$ & 0.65 $\pm 0.32_{17}$ & 0.01 $\pm 0.00_{2}$ & 0.48 $\pm 0.37_{10}$ & 0.52 $\pm 0.42_{10}$ \\
Ionosphere & - & - & - & - & - & - & 21.3 $\pm$ 8.2 & 6776.0 $\pm$ 3324.5 & 0.64 $\pm 0.41_{23}$ & 0.82 $\pm 0.32_{14}$ & 0.25 $\pm 0.20_{5}$ & 0.76 $\pm 0.23_{5}$ & 0.81 $\pm 0.16_{3}$ & 0.59 $\pm 0.41_{2}$ \\
Quake & - & - & - & - & - & - & 11.6 $\pm$ 4.8 & 9585.9 $\pm$ 899.3 & 0.91 $\pm 0.20_{19}$ & 0.87 $\pm 0.24_{18}$ & 0.64 $\pm 0.40_{15}$ & - & 0.68 $\pm 0.36_{13}$ & 0.71 $\pm 0.33_{16}$ \\
Wine & - & - & - & - & - & - & 14.4 $\pm$ 7.3 & 8685.9 $\pm$ 2585.8 & 0.82 $\pm 0.31_{24}$ & 0.86 $\pm 0.24_{21}$ & 0.69 $\pm 0.30_{18}$ & 0.33 $\pm 0.29_{2}$ & 0.53 $\pm 0.38_{17}$ & 0.74 $\pm 0.31_{13}$ \\
\cline{8-15}
&&&&&&&16.15 2.86 & 8447.63 $\pm $1170.97 & 0.75 $\pm 0.09_{21.80 \pm 3.92}$ & 0.82 $\pm 0.06_{18.40 \pm 3.56}$ & 0.65 $\pm 0.19_{12.90 \pm 4.41}$ & 0.47 $\pm 0.32_{5.75 \pm 4.92}$ & 0.67 $\pm 0.11_{15.11 \pm 5.40}$ & 0.68 $\pm 0.09_{10.50 \pm 4.92}$ \\
\hline
\end{tabular}}
\end{table*}

\begin{table*}[htb]
\caption{Results for outer algorithm DE with ARM metric weight adaptation and selecting more preprocessing methods.}
\label{tab:res_multi_preproc_ow_DE}
\centering
\resizebox{\textwidth}{!}{
\begin{tabular}{c|cccccc|c|c|cccccc}
\hline
\multirow{2}{*}{Dataset} & \multicolumn{6}{c|}{Preprocessing method} & \multicolumn{2}{c|}{Hyper-parameters} & \multicolumn{6}{c|}{Metrics \& Weights}\\ \cline{2-15} 
& $MM$ & $ZS$ & $DS$ & $RHC$ & $KM$ & $N^{a}$ & $\mathit{NP}$ & $\mathit{MAXFES}$ & Supp & Conf & Cover & Amp & Incl & Comp\\ \hline
Abalone & - & - & - & - & - & - & 11.6 $\pm$ 4.2 & 8989.0 $\pm$ 1818.6 & 0.73 $\pm 0.23_{25}$ & 0.74 $\pm 0.29_{17}$ & 0.85 $\pm 0.23_{15}$ & - & 0.72 $\pm 0.30_{21}$ & 0.67 $\pm 0.40_{9}$ \\
Balance scale & - & - & - & - & - & - & 15.0 $\pm$ 6.7 & 7358.0 $\pm$ 3076.0 & 0.61 $\pm 0.31_{24}$ & 0.74 $\pm 0.32_{24}$ & 0.44 $\pm 0.32_{12}$ & - & 0.82 $\pm 0.27_{16}$ & 0.60 $\pm 0.29_{10}$ \\
Basketball & - & - & - & - & - & - & 13.6 $\pm$ 5.5 & 8971.7 $\pm$ 1704.6 & 0.69 $\pm 0.27_{25}$ & 0.72 $\pm 0.33_{19}$ & 0.57 $\pm 0.33_{13}$ & - & 0.74 $\pm 0.31_{20}$ & 0.61 $\pm 0.37_{15}$ \\
Bolts & - & - & - & - & - & - & 15.6 $\pm$ 6.5 & 8468.6 $\pm$ 2388.3 & 0.73 $\pm 0.27_{21}$ & 0.76 $\pm 0.28_{20}$ & 0.71 $\pm 0.35_{17}$ & 0.34 $\pm 0.17_{3}$ & 0.81 $\pm 0.23_{26}$ & 0.63 $\pm 0.36_{10}$ \\
Buying & - & - & - & - & - & -& 15.1 $\pm$ 6.3 & 9024.1 $\pm$ 1431.3 & 0.72 $\pm 0.32_{30}$ & 0.61 $\pm 0.32_{12}$ & 0.67 $\pm 0.33_{2}$ & - & - & - \\
German & - & - & - & - & - & - & 22.2 $\pm$ 7.6 & 6033.7 $\pm$ 2926.3 & 0.55 $\pm 0.33_{11}$ & 0.62 $\pm 0.33_{22}$ & 0.40 $\pm 0.35_{14}$ & 0.57 $\pm 0.32_{15}$ & 0.59 $\pm 0.31_{13}$ & 0.72 $\pm 0.32_{11}$ \\
House16 & - & - & - & - & - & -& 15.8 $\pm$ 7.3 & 7880.9 $\pm$ 2238.0 & 0.77 $\pm 0.29_{25}$ & 0.74 $\pm 0.28_{23}$ & 0.68 $\pm 0.29_{21}$ & - & 0.55 $\pm 0.36_{13}$ & 0.54 $\pm 0.38_{14}$ \\
Ionosphere & - & - & - & - & - & - & 16.8 $\pm$ 7.3 & 8059.6 $\pm$ 2564.3 & 0.71 $\pm 0.34_{28}$ & 0.82 $\pm 0.28_{21}$ & 0.52 $\pm 0.36_{5}$ & - & 0.53 $\pm 0.39_{5}$ & - \\
Quake & - & - & - & - & - & - & 11.8 $\pm$ 2.8 & 8982.1 $\pm$ 1247.7 & 0.78 $\pm 0.29_{27}$ & 0.66 $\pm 0.30_{15}$ & 0.73 $\pm 0.28_{13}$ & 0.21 $\pm 0.00_{1}$ & 0.64 $\pm 0.36_{18}$ & 0.71 $\pm 0.30_{18}$ \\
Wine & - & - & - & - & - & - & 14.6 $\pm$ 5.9 & 9342.5 $\pm$ 1265.8 & 0.65 $\pm 0.34_{24}$ & 0.83 $\pm 0.24_{29}$ & 0.66 $\pm 0.33_{17}$ & 0.08 $\pm 0.00_{1}$ & 0.63 $\pm 0.33_{22}$ & 0.67 $\pm 0.32_{11}$ \\
\cline{8-15}
&&&&&&&15.22 $\pm$ 2.82 & 8311.03 $\pm$ 963.66 & 0.69 $\pm 0.07_{24.00 \pm 4.92}$ & 0.72 $\pm 0.07_{20.20 \pm 4.58}$ & 0.62 $\pm 0.13_{12.90 \pm 5.36}$ & 0.30 $\pm 0.18_{5.00 \pm 5.83}$ & 0.67 $\pm 0.10_{17.11 \pm 5.86}$ & 0.64 $\pm 0.05_{12.25 \pm 2.90}$\\
\hline
\end{tabular}}
\end{table*}

\subsubsection{Comparison with VARDE state-of-the-art algorithm}
The last experiment was reserved for an indirect comparison with the VARDE state-of-the-art algorithm~\cite{mlakar2023variable} for ARM, which represents a hybridized version of DE, that was designed specifically for the exploration and exploitation of the ARM search space. Thus, the best reported variations of VARDE were used in this comparative study. It is not a direct comparison, since the pipelines produced by NiaAutoARM are dataset specific. Therefore, for each dataset, we observed which components of the pipeline provided the best results (i.e., the inner algorithm, preprocessing component and rule evaluation metrics), and performed 30 independent runs with these settings. The results of these dataset specific independent runs were compared to the results of VARDE by using the Wilcoxon signed rank test. 

The results are depicted in Table~\ref{tab:wilcox_varde}. 
\begin{table}[htb]
    \caption{Results of the Wilxocon test, comparing the NiaAutoARM generated pipelines with VARDE.}
    \label{tab:wilcox_varde}
    \centering
    \resizebox{0.5\textwidth}{!}{
    \begin{tabular}{c|cc|cc|cc|c|}
    \hline
        \multirow{2}{*}{} & \multicolumn{2}{c|}{Baseline} & \multicolumn{2}{c|}{WO, $P=1$} & \multicolumn{2}{c|}{WO, $P>1$} \\ \cline{2-7}
        &PSO&DE&PSO&DE&PSO&DE\\ \hline
        VARDE\_pos\_15\_2000\cite{mlakar2023variable} &\textbf{0.03} & 0.34& \textbf{0.01}&0.08&\textbf{0.01}&\textbf{0.01}\\ 
        VARDE\_neg\_15\_2000~\cite{mlakar2023variable} &0.61 &0.17&0.97&0.54 &0.75 &0.98 \\
        \hline
    \end{tabular}}
\end{table}
As is evident from the Table, the pipelines found by the NiaAutoARM provided significantly better results in some instances compared to the VARDE method. Therefore, the NiaAutoARM is distinguished as an effective framework for ARM.

\begin{table}[htb]
    \caption{Average execution times of both algorithms, needed for finding the best pipelines for each experimental dataset in seconds.}
    \label{tab:execution_time}
    \centering
    \begin{tabular}{c|l l}
    \hline
       Dataset & PSO & DE \\
       \hline
       Abalone       & $27584   \pm 7238.7$   &$23486.5 \pm 4702.6$\\
       Balance scale & $15356.1 \pm 6617$      &$11598.7 \pm 1298.3$\\
       Basketball    & $23442.6 \pm 5271.6$   &$15476.7 \pm 1893.6$\\
       Bolts         & $22325.9 \pm 9694.9$   &$18603.7 \pm 4979.5$\\
       Buying        & $33819.2 \pm 10046$     &$34449.2 \pm 4134.3$\\
       German        & $25322.6 \pm 10027.3$   &$25958.7 \pm 3230.3$\\
       House         & $34444.4 \pm 8286.6$    &$34464   \pm 7709.4$\\
       Ionosphere    & $32299.7 \pm 9396.3$   &$40831.1 \pm 7365.6$\\
       Quake         & $17897.9 \pm 4523.5$   &$18393.1 \pm 4162.1$\\
       Wine          & $28541.7 \pm 7341.7$   &$24963.4 \pm 3111.6$\\
       \hline
    \end{tabular}
\end{table}

\subsection{Discussion}\label{sec:discussion}
The results show notable trends in the optimization of ARM pipelines. The PSO algorithm was selected predominantly over jDE, DE, LSHADE, and ILSHADE as the inner optimization method. This preference can be attributed to the PSO's ability to balance exploration and exploitation effectively, enabling it to navigate the search space efficiently and avoid premature convergence. In contrast, the other algorithms may converge too quickly, potentially limiting their effectiveness in identifying diverse high-quality pipelines, and making them less suitable for this specific optimization task. Min-max scaling was the most frequently used preprocessing method, likely due to its simplicity and ability to standardize data efficiently. Additionally, support and confidence were the dominant metrics in the generated pipelines, reflecting their fundamental role in ARM.

While the approach exhibits a slightly higher computational complexity due to the iterative optimization and exploration of diverse preprocessing combinations, this is a manageable trade-off (see Table~\ref{tab:execution_time}). The superior results achieved, particularly in comparison to the VARDE state-of-the-art hybrid DE method, underscore the robustness of the approach. Notably, the method operates without requiring prior knowledge of the algorithms or datasets, making it adaptable and versatile for various applications.

\section{Conclusion}\label{sec:conclusion}

This paper presented NiaAutoARM, an innovative framework designed for the optimization of the ARM pipelines using stochastic population-based Nia-s. The framework integrates the selection of: an inner ARM heuristic, its hyper-parameter optimization, dataset preprocessing techniques, and searching for the more suitable fitness function represented as a weighted sum of ARM evaluation metrics, where the weights are subjects of the adaptation. Extensive evaluations on ten widely used datasets from the UC Irvine repository underscore the framework’s effectiveness, particularly for users with limited domain expertise. Comparative analysis against the VARDE state-of-the-art hybrid DE highlights the superior performance of the proposed framework in generating high-quality ARM pipelines further.

The future work would aims to address several key areas: First, integrating additional Nia-s with adaptive parameter tuning could enhance the pipeline optimization process further. Second, incorporating other advanced preprocessing techniques and alternative metrics might improve pipeline diversity and domain-specific applicability. Third, exploring parallel and distributed computing strategies could mitigate computational complexity, making the framework more scalable. Finally, extending the framework to support multi-objective optimization would enable a deeper exploration of trade-offs between conflicting metrics, advancing its utility further in real-world applications.

\ifCLASSOPTIONcompsoc
  \section*{Acknowledgments}
\else
  \section*{Acknowledgment}
\fi

Iztok Fister Jr. thanks the financial support from the Slovenian Research Agency (Research Core Funding No. P2-0057). Uroš Mlakar thanks the financial support from the Slovenian Research and Innovation Agency (Program No. P2-0041). The authors express their gratitude to Žiga Stupan for his insightful input during the initial discussions of this research.

\bibliographystyle{IEEEtran}
\bibliography{bibtex.bib}

\end{document}